# Quaternion variational integration for inertial maneuvering in a biomimetic UAV

**Arion Pons**[1]
Department of Engineering, University of Cambridge
Cambridge CB2 1PZ, UK

Department of Mechanics and Maritime Sciences, Chalmers University of Technology
Gothenburg 412 96, Sweden

**Fehmi Cirak**
Department of Engineering, University of Cambridge
Cambridge CB2 1PZ, UK

**ABSTRACT**

*Biological flying, gliding, and falling creatures are capable of extraordinary forms of inertial maneuvering: free-space maneuvering based on fine control of their multibody dynamics, as typified by the self-righting reflexes of cats. However, designing inertial maneuvering capability into biomimetic robots, such as biomimetic unmanned aerial vehicles (UAVs) is challenging. Accurately simulating this maneuvering requires numerical integrators that can ensure both singularity-free integration, and momentum and energy conservation, in a strongly coupled system – properties unavailable in existing conventional integrators. In this work, we develop a pair of novel quaternion variational integrators (QVIs) showing these properties, and demonstrate their capability for simulating inertial maneuvering in a biomimetic UAV showing complex multibody-dynamics coupling. Being quaternion-valued, these QVIs are innately singularity-free; and being variational, they can show excellent energy and momentum conservation properties. We explore the effect of variational integration order (left-rectangle vs. midpoint) on the conservation properties of integrator, and conclude that, in complex coupled systems in which canonical momenta may be time-varying, the midpoint integrator is required. The resulting midpoint QVI is well-suited to the analysis of inertial maneuvering in a biomimetic UAV – a feature that we demonstrate in simulation – and of other complex dynamical systems.*

---

[1] Corresponding author. Email: arion@chalmers.se

# 1 INTRODUCTION

Inertial maneuvering is a form of free-space orientation control that is common in biological creatures: it involves the use of biological multibody dynamics (*e.g.*, the effects of wing or limb inertia) to achieve orientation changes independent of aerodynamic forces. As a free-fall response, it is found in a wide range of flightless creatures – in cats, where it is responsible for their extraordinary capability to land on their feet [1]; as well as geckos [2] and flightless insects [3]. As a flight control mechanism, it is observed in a range of flying and gliding creatures – enabling complex vertical landing maneuvers [4], low-airspeed turns [5], and limbless flight control [6]. Both in-flight and flightless inertial maneuvering are relevant to the design of biomimetic robots: to self-stabilizing biomimetic terrestrial robots [7–9], and to biomimetic uncrewed aerial vehicles (UAVs) [10,11].

Simulating inertial maneuvering in such systems poses several challenges, among which is the problem of singularity-free simulation. For a system which may mass through any orientation – for instance, a UAV carrying out a vertical landing maneuver – the orientation representation used must be free from gimbal lock [12,13]. One approach to ensuring singularity-free simulation involves the quaternions: four-parameter hypercomplex numbers [14]. The set of unit quaternions is a Lie group [15,16], providing a smooth and compact method of representing orientation. They allow singularity-free analysis of the dynamics of aircraft [17], spacecraft [18], biological and robotic manipulators [19,20], and many other structures. However, unit quaternions are not well-suited to conventional numerical integration techniques: the constraint to the unit hypersphere necessitates that numerical integration be carried out not in unconstrained space, but on this hypersphere. Neglect of this property can lead to a degradation of integrator quality and generality [21–23].

This degradation has spurred the development of dedicated quaternion numerical integrators. Crouch-Grossman [16,24] and Runge-Kutta-Muthe-Kaas [25,26] methods represent applications of the traditional high-order Runge-Kutta methods to the quaternion unit hypersphere. In addition, quaternion orientation representation can be integrated within variational integration methods, leading to quaternion variational integrators (QVIs) [15,18,27]; a subclass of Lie group variational integrator [28,29]. Variational integrators represent a recent development in computational mechanics, showing several advantages over non-variational integration: in particular, excellent energy and momentum conservation properties [30]. Taken together, quaternion orientation representation and variational integration represent an ideal pairing for the simulation of inertial maneuvering in biomimetic and biological systems: a pairing that provides a highly accurate characterization of the kinetic energy and momentum transfers involved in inertial maneuvering, over a singularity-free orientation space. However, existing QVIs are not directly applicable to the complex dynamics of multibody inertial maneuvering systems – instead, being focused on uncoupled single-body motion, orbital mechanics, and beam dynamics [15,18,27]. Biomimetic UAVs undergoing inertial maneuvering show complex and strongly-coupled dynamics [11]: despite ongoing efforts, there remain no QVIs available for these systems.

In this work, we seek to remedy this deficit, with the development of a pair of QVIs – a left-rectangle, and a midpoint integrator – for a complex and general multibody-dynamic model of inertial maneuvering in a biomimetic UAV. We find that retaining favorable energy and momentum conservation in this strongly-coupled system requires midpoint integration. The application of quaternion variational integration to this biomimetic UAV relieves an analysis bottleneck associated with inertial maneuvering simulation availability and accuracy; and the resulting integrators are applicable not only to analogous models, *e.g.*, of biological creatures; but to other strongly-coupled systems.

## 2 DYNAMIC MODEL OF INERTIAL MANEUVERING

### 2.1. Multibody modelling of a biomimetic UAV

Pons and Cirak [10,11] developed a flight dynamic model of complex maneuverability, including inertial maneuvering, in a biomimetic morphing-wing UAV. Fig. 1 shows this UAV, alongside one associated form of inertial maneuvering: inverted take-off and landing, from a fixed inverted landing state, to an arbitrary heading. The dynamic model of this UAV can be expressed in Lagrangian form, via the UAV kinetic energy ($T$) [10,11]:

$$T(\dot{\mathbf{x}}^{(b)}, \boldsymbol{\omega}^{(b)}, \mathbf{v}) = \dot{\mathbf{x}}^{(b),T} \mathrm{a}_{xx} \dot{\mathbf{x}}^{(b)} + \dot{\mathbf{x}}^{(b),T} \mathrm{A}_{x\omega}(\mathbf{v}) \boldsymbol{\omega}^{(b)} + \dot{\mathbf{x}}^{(b),T} \mathbf{a}_x(\mathbf{v}) \\ + \boldsymbol{\omega}^{(b),T} \mathrm{A}_{\omega\omega}(\mathbf{v}) \boldsymbol{\omega}^{(b)} + \boldsymbol{\omega}^{(b),T} \mathbf{a}_\omega(\mathbf{v}) + \mathrm{a}_0(\mathbf{v}),$$

(1)

where $\dot{\mathbf{x}}^{(b)}$ and $\boldsymbol{\omega}^{(b)}$ are the UAV translational and angular velocities, both in the body-fixed reference frame; $\mathbf{v}$ is a vector of morphing parameters; and $\mathrm{a}_{xx}$, $\mathrm{a}_0$, $\mathrm{A}_{x\omega}$, $\mathrm{A}_{\omega\omega}$, $\mathbf{a}_x$, and $\mathbf{a}_\omega$ are coefficient functions, given in [10,11]. We assume these coefficients are continuous functions of $\mathbf{v}$, but do not otherwise prescribe their form. The analysis can also be extended to include a gravitational potential; but here we focus on the kinetic energy, $T$, for three reasons: firstly, the kinetic energy is more complex, whereas a gravitational potential is a comparatively simple addition; secondly, because, gravitational forces necessarily do not directly contribute to overall reorientation; and thirdly because gravity-free analyses are used in the analysis of biological inertial maneuvering [1,2,4], and we will seek to imitate some of these biological maneuvers.

Eq. 1 is well-suited for variational integration, because the coefficients $a_{xx}$, *etc.*, are independent of the state variables, $\dot{\mathbf{x}}^{(b)}$ and $\boldsymbol{\omega}^{(b)}$ – simplifying the variational analysis. The orientation of the UAV is represented with a quaternion, $q$, on the unit hypersphere, $\|q\| = 1$. This quaternion defines the relationship between the UAV body-fixed reference frame, denoted $(\cdot)^{(b)}$, and the earth reference frame, denoted $(\cdot)^{(e)}$. Under the Hamilton convention [31], this relationship may be expressed:

$$\begin{aligned} \mathbf{x}^{(e)} &= q \otimes \mathbf{x}^{(b)} \otimes q^\dagger, \\ \mathbf{x}^{(b)} &= q^\dagger \otimes \mathbf{x}^{(e)} \otimes q, \end{aligned} \quad (2)$$

for any 3-vector, $\mathbf{x}$, resolved in either frame. $\otimes$ and $(\cdot)^\dagger$ denote quaternion multiplication and transpose, respectively – see [14] for further properties of quaternion algebra. Note that $\mathbf{x}$ functions as a purely imaginary quaternion, and is thus subject to quaternion operators. The dynamic behavior of the UAV can be expressed via the Hamilton's principle, asserting that the system action functional is stationary with respect to first-order perturbations [32]. For inertial maneuvering without a gravitational potential, we may express the UAV dynamics as:

$$\int_{t_0}^{t_n} \delta T + \mathbf{Q} \cdot \delta \mathbf{r} \, dt = 0, \quad (3)$$

for variational derivative of kinetic energy $\delta T$, generalized forces $\mathbf{Q}$, generalized coordinates $\mathbf{r}$, time $t$, and simulation time interval $[t_0, t_n]$.

**2.2. Choice of dimensional-reduction map**

To translate Eq. 3 into a numerical integrator, Eq. 1-3, a set of generalized coordinates and associated time derivatives are required. For a quaternion generalized coordinate $q$, direct time differentiation is unsuitable: the derivative $\dot{q}$ is an underconstrained parameterization of the orientation rate, requiring an additional constraint in the integrator inter-step equations derived from the unit hypersphere constraint. In addition, if $q$-$\dot{q}$ is selected as a pairing, then the angular velocity, $\boldsymbol{\omega}^{(b)}$, as used in Eq. 1, is given by:

$$\boldsymbol{\omega}^{(b)} = 2q^\dagger \otimes \dot{q}. \quad (4)$$

The kinetic energy, $T$, then has a dependency on generalized coordinate, $q$, which complicates the variational analysis by introducing further terms in $\delta T$.

Recent variational treatments of the constraint on $\dot{q}$ utilize the following three-step process [15,18]. (**1**) In a discrete or infinitesimal continuous framework, the rotation of $q$ on the hypersphere between adjacent time-steps will be small. (**2**) Under a small-rotation approximation, these rotations can be parameterized by a local dimensional-reduction map (diffeomorphism) generating a local unconstrained 3-parameter representation. (**3**) Local discrete or infinitesimal rotations can be solved in this local unconstrained representation, and then mapped back to full quaternion space. (**end**). If we denote $q_{k+1}$ and $q_k$ as unit quaternions at adjacent discrete time-steps; $q = M(\boldsymbol{\theta})$ as a diffeomorphism to $\boldsymbol{\theta} \in \mathbb{R}^3$; and $F(\cdot)$ as the solution of the dynamical system; then this three-step process may be expressed:

(**1**) $\quad q_{k+1} = q_k \otimes \Delta q_k, \quad \Delta q_k \to 1;$

(**2**) $\quad \boldsymbol{\theta}_{k+1} = M^{-1}(q_{k+1}), \quad \boldsymbol{\theta}_k = M^{-1}(q_k), \quad \Delta\boldsymbol{\theta}_k = M^{-1}(\Delta q_k);$ (5)

(**3**) $\quad \Delta\boldsymbol{\theta}_k = F(\boldsymbol{\theta}_k, \ldots), \quad q_{k+1} = q_k \otimes M(\Delta\boldsymbol{\theta}_k).$

Two diffeomorphisms, $M(\boldsymbol{\theta})$, are known to be suitable for this process: the exponential map, and the Cayley map [15]. Existing QVIs have commonly used the Cayley map [15,18]:

$$M(\boldsymbol{\theta}) = \frac{1 + \boldsymbol{\theta}}{\sqrt{1 + \|\boldsymbol{\theta}\|^2}}, \tag{6}$$

However, as an alternative for systems with complex dynamics, such as our inertial maneuvering UAV, we utilize the exponential map:

$$M(\boldsymbol{\theta}) = \exp(\boldsymbol{\theta}). \tag{7}$$

for quaternion exponential $\exp(\cdot)$ [31]. An advantage of this approach is that the physical angular velocity $\boldsymbol{\omega}^{(b)} = d\boldsymbol{\theta}/dt$ (a logarithmic derivative) becomes the proxy derivative for the orientation quaternion. The continuous relation between $\dot{q}$ and $\boldsymbol{\omega}^{(b)}$ is given directly by Eq. 3; and the discrete relation can be estimated by any quaternion kinematic integrator, as per [16,26]. We utilize the first-order Crouch-Grossman (CG) method, with step size $h$:

$$q_{k+1} = q_k \otimes \exp\left(\frac{1}{2}h\boldsymbol{\omega}_k^{(b)}\right), \tag{8}$$

## 3 INTEGRATOR FORMULATIONS

### 3.1. Left-rectangle integrator

In a discrete mechanics framework [30], the time integral in Hamilton's principle, Eq. 3, may be approximated via left-rectangle integration:

$$\delta h \sum_{k=1}^{N-1} \delta T_k + \mathbf{Q}_k \cdot \delta \mathbf{r}_k = 0. \tag{9}$$

The variational derivative of $T_k$, $\delta T_k$, is defined with reference to first-order perturbations in orientation ($q_k$) and position ($\mathbf{x}^{(e)}$). Computing $\delta T_k$ involves defining these perturbations, and propagating them through other kinematic variables ($\dot{\mathbf{x}}^{(b)}$ and $\boldsymbol{\omega}^{(b)}$) to $T_k$. First, defining the perturbations. Following [18], the orientation, $q_k$, is subjected to a continuous norm-preserving perturbation defined via the quaternion exponential:

$$\begin{aligned} q_k^\epsilon &= q_k \otimes \exp\left(\epsilon \boldsymbol{\eta}_k^{(b)}\right) \cong q_k + \delta q_k + \mathcal{O}(\epsilon^2) \\ &= q_k + \epsilon q_k \otimes \boldsymbol{\eta}_k^{(b)} + \mathcal{O}(\epsilon^2), \end{aligned} \tag{10}$$

where $\boldsymbol{\eta}_k^{(b)}$ represents a perturbative angular velocity axis in the body-fixed frame – the axis around which the system will be perturbed by a small angle. This perturbation can be propagated to $\boldsymbol{\omega}^{(b)}$ via a first-order approximation, of Eq. 4:

$$\begin{aligned} \boldsymbol{\omega}_k^{(b)} &= \frac{2}{h} q_k^\dagger \otimes (q_{k+1} - q_k), \\ \delta \boldsymbol{\omega}_k^{(b)} &= \boldsymbol{\omega}_k^{(b)} \otimes \boldsymbol{\eta}_{k+1}^{(b)} - \boldsymbol{\eta}_k^{(b)} \otimes \boldsymbol{\omega}_k^{(b)} + \frac{2}{h}\left(\boldsymbol{\eta}_{k+1}^{(b)} - \boldsymbol{\eta}_k^{(b)}\right). \end{aligned} \tag{11}$$

For position, $\mathbf{x}^{(e)}$, perturbation is more straightforward, as there are no constraints. We utilize:

$$\mathbf{x}_k^{(e),\epsilon} = \mathbf{x}_k^{(e)} + \epsilon \delta \mathbf{x}_k^{(e)} + \mathcal{O}(\epsilon^2). \tag{12}$$

Together with the angular velocity perturbation of Eq. 11, this yields the perturbation on $\dot{\mathbf{x}}_k^{(b)}$:

$$\delta \dot{\mathbf{x}}_k^{(b)} = 2\dot{\mathbf{x}}_k^{(b)} \times \boldsymbol{\eta}_k^{(b)} + \frac{1}{h} q_k^\dagger \left(\delta \mathbf{x}_{k+1}^{(e)} - \delta \mathbf{x}_k^{(e)}\right) q_k. \tag{13}$$

Finally, for the complete system of Eq. 1, the variational derivative of the Lagrangian, $\delta T_k$, can be expressed via the chain rule as:

$$\begin{aligned} \delta T_k &= \mathbf{D}_{1,k} \cdot \delta \dot{\mathbf{x}}_k^{(b)} + \mathbf{D}_{2,k} \cdot \delta \boldsymbol{\omega}_k^{(b)}, \\ \mathbf{D}_{1,k}\left(\dot{\mathbf{x}}_k^{(b)}, \boldsymbol{\omega}_k^{(b)}, \mathbf{v}_k\right) &= \partial T_k / \partial \dot{\mathbf{x}}_k^{(b)} = 2\mathrm{a}_{xx}(\mathbf{v}_k)\dot{\mathbf{x}}_k^{(b)} + \mathrm{A}_{x\omega}(\mathbf{v}_k)\boldsymbol{\omega}_k^{(b)} + \mathbf{a}_x(\mathbf{v}_k), \\ \mathbf{D}_{2,k}\left(\dot{\mathbf{x}}_k^{(b)}, \boldsymbol{\omega}_k^{(b)}, \mathbf{v}_k\right) &= \partial T_k / \partial \boldsymbol{\omega}_k^{(b)} = 2\mathrm{A}_{\omega\omega}(\mathbf{v}_k)\boldsymbol{\omega}_k^{(b)} + \mathrm{A}_{x\omega}(\mathbf{v}_k)^T \dot{\mathbf{x}}_k^{(b)} + \mathbf{a}_\omega(\mathbf{v}_k). \end{aligned} \tag{14}$$

The discretized Hamilton's principle, Eq. 9, is then directly expressible as two coupled inter-step equations; in translation and rotation, coupling step $k-1$ to step $k$:

$$\begin{aligned}
q_k \mathbf{D}_{1,k} q_k^\dagger &= q_{k-1} \mathbf{D}_{1,k-1} q_{k-1}^\dagger + h \mathbf{F}_k^{(e)}, \\
\mathbf{D}_{2,k} + \frac{1}{2} h \boldsymbol{\omega}_k^{(b)} \times \mathbf{D}_{2,k} + h \dot{\mathbf{x}}_k^{(b)} \times \mathbf{D}_{1,k} &= \mathbf{D}_{2,k-1} - \frac{1}{2} h \boldsymbol{\omega}_{k-1}^{(b)} \times \mathbf{D}_{2,k-1} + \boldsymbol{\tau}_k^{(b)}.
\end{aligned} \quad (15)$$

In this form, the generalized force, $\mathbf{Q}$, may be split into translational force $\mathbf{F}^{(e)}$, in the earth frame, and rotational moment $\boldsymbol{\tau}^{(b)}$, in the body-fixed frame.

It is worth observing these inter-step equations more closely. The equations take the form of momentum balances in the system's generalized, or canonical momenta [33]; here:

$$\begin{aligned}
\mathbf{p}_{x,k} &= q_k \mathbf{D}_{1,k} q_k^\dagger, \\
\mathbf{p}_{\omega,k} &= \mathbf{D}_{2,k,i} + \frac{1}{2} h \boldsymbol{\omega}_k^{(b)} \times \mathbf{D}_{2,k,i}.
\end{aligned} \quad (16)$$

The translational equation is symmetric, with momentum term $\mathbf{p}_{x,k}$. In the absence of external force, $\mathbf{F}^{(e)}$, Eq. 15 ensures that the translational momentum of the UAV is conserved in a numerical-integration context – a favorable property of the flight simulation. The rotational equation is nearly symmetric, with momentum term $\mathbf{p}_{\omega,k}$: the exception is the antisymmetric term, $h \dot{\mathbf{x}}_k^{(b)} \times \mathbf{D}_{1,k}$. The presence of this antisymmetric term allows for potentially problematic rotational momentum drift in simulation, an cannot simply be eliminated via an alternative generalized coordinate formulation (*e.g.*, with $\dot{\mathbf{x}}_k^{(e)}$, $\boldsymbol{\omega}_k^{(e)}$). The indirect dependency of the system kinetic energy on generalized coordinates – contained in the proxy derivative relation, Eq. 11 – means that at least one of the system's canonical momenta will always not be conserved, as the associated generalized coordinate will not be ignorable [33]. Finally, as is the case even in uncoupled dynamical systems [18], this inter-step equation, Eq. 15, does not permit an analytical solution, but must instead be solved numerically – we utilize Newton's iteration, starting from the previous step. This completes the derivation of the left-rectangle QVI.

**3.2. Midpoint integrator**

The implication of the canonical-momenta analysis in Section 3.1 is that left-rectangle QVIs, while suitable for uncoupled rigid-body dynamics [18]; are likely to be unsuitable for the complex coupling present in an inertial maneuvering simulation. A more accurate form of integration is thus one key avenue to re-attaining good conservation properties in this situation.

Here we apply midpoint integration, which we would expect to capture linear trends in $T$. Applying discrete midpoint integration to Hamilton's principle, Eq. 2, we obtain:

$$h \sum_{k=1}^{N-1} \delta \widetilde{T}_k + \widetilde{\mathbf{Q}}_k \cdot \delta \widetilde{\mathbf{r}}_k = 0, \tag{17}$$

where the tilde ($\tilde{x}$) denotes evaluation at an inter-step midpoint. The midpoint location in non-quaternion and imaginary quaternion variables ($t$, $\mathbf{x}^{(e)}$, *etc.*) can be computed via linear interpolation. However, to compute the midpoint quaternion ($\tilde{q}_k$) a different scheme is necessary, as linear interpolation does not preserve the orientation quaternion normalization [34]. Other schemes available include normalized linear interpolation (NLERP), spherical linear interpolation (SLERP), spherical spline interpolation (SQUAD), and other methods [34,35]. For any choice of interpolation, the variational derivative of $\widetilde{T}_k$ is:

$$\begin{aligned}
\delta \widetilde{T}_k &= \widetilde{\mathbf{D}}_{1,k} \cdot \delta \widetilde{\mathbf{x}}_k^{(b)} + \widetilde{\mathbf{D}}_{2,k} \cdot \delta \widetilde{\boldsymbol{\omega}}_k^{(b)}, \\
\mathbf{D}_{1,k}\left(\widetilde{\mathbf{x}}_k^{(b)}, \widetilde{\boldsymbol{\omega}}_k^{(b)}, \tilde{\mathbf{v}}_k\right) &= \partial \widetilde{T}_k / \partial \widetilde{\mathbf{x}}_k^{(b)} = 2\mathrm{a}_{xx}\widetilde{\mathbf{x}}_k^{(b)} + \mathrm{A}_{x\omega}(\tilde{\mathbf{v}}_k)\boldsymbol{\omega}_k^{(b)} + \mathbf{a}_x(\tilde{\mathbf{v}}_k), \\
\mathbf{D}_{2,k}\left(\widetilde{\mathbf{x}}_k^{(b)}, \widetilde{\boldsymbol{\omega}}_k^{(b)}, \tilde{\mathbf{v}}_k\right) &= \partial \widetilde{T}_k / \partial \widetilde{\boldsymbol{\omega}}_k^{(b)} = 2\mathrm{A}_{\omega\omega}(\tilde{\mathbf{v}}_k)\boldsymbol{\omega}_k^{(b)} + \mathrm{A}_{x\omega}(\tilde{t}_k)^T \widetilde{\mathbf{x}}_k^{(b)} + \mathbf{a}_\omega(\tilde{t}_k).
\end{aligned} \tag{18}$$

The variational derivatives of the proxy derivatives at the midpoint, $\delta \widetilde{\mathbf{x}}_k^{(b)}$ and $\delta \widetilde{\boldsymbol{\omega}}_k^{(b)}$, must then be related to the perturbations at the step points ($k$ and $k+1$). The midpoint proxy derivatives and their perturbations are:

$$\begin{aligned}
\widetilde{\boldsymbol{\omega}}_k^{(b)} &= 2\tilde{q}_k^\dagger \dot{\tilde{q}}_k, & \widetilde{\boldsymbol{\omega}}_k^{(b),\epsilon} &= 2\tilde{q}_k^{\epsilon,\dagger} \dot{\tilde{q}}_k^\epsilon, \\
\widetilde{\mathbf{x}}_k^{(b)} &= \tilde{q}_k^\dagger \tilde{\mathbf{x}}^{(e)} \tilde{q}_k, & \widetilde{\mathbf{x}}_k^{(b),\epsilon} &= \tilde{q}_k^{\epsilon,\dagger} \tilde{\mathbf{x}}^{(e),\epsilon} \tilde{q}_k^\epsilon.
\end{aligned} \tag{19}$$

Here the central obstacle of the midpoint integrator arises, which is the definition of the perturbed midpoint, $\tilde{q}_k^\epsilon$. The unperturbed midpoint can be computed the interpolation methods noted earlier – for instance, under SLERP and NLERP [34]:

$$\begin{aligned}
\text{SLERP:} \qquad & \tilde{q}_k = \left(q_{k+1} q_k^\dagger\right)^{1/2} q_k, \\
\text{NLERP:} \qquad & \tilde{q}_k = \frac{q_{k+1} + q_k}{\|q_{k+1} + q_k\|}.
\end{aligned} \tag{20}$$

Perturbations of these quaternion interpolation functions are not easy to compute in closed form. To generate the best approximation, we diverge from the analysis of Section 3.1 and define perturbations about a direction in the earth (not, body-fixed) reference frame:

$$q_k^\epsilon = \exp\left(\epsilon \boldsymbol{\eta}_k^{(e)}\right) q_k \cong q_k + \epsilon \boldsymbol{\eta}_k^{(e)} q_k + \mathcal{O}(\epsilon^2). \tag{21}$$

This alteration simplifies the manipulation, and is not fundamental – it may be verified that the equivalence between the earth and body perturbations is exact:

$$q_k^\epsilon = \exp\left(\epsilon \boldsymbol{\eta}_k^{(e)}\right) q_k = \exp\left(\epsilon q_k \boldsymbol{\eta}_k^{(b)} q_k^\dagger\right) q_k = q_k \exp\left(\epsilon \boldsymbol{\eta}_k^{(b)}\right). \tag{22}$$

By parameterizing the perturbed midpoint in a local perturbative direction $\widetilde{\boldsymbol{\eta}}_k^{(e)}$:

$$\begin{aligned}\tilde{q}_k^\epsilon &= \exp\left(\epsilon \widetilde{\boldsymbol{\eta}}_k^{(e)}\right) \tilde{q}_k \cong \tilde{q}_k + \epsilon \widetilde{\boldsymbol{\eta}}_k^{(e)} \tilde{q}_k + \mathcal{O}(\epsilon^2), \\ \dot{\tilde{q}}_k^\epsilon &\cong \dot{\tilde{q}}_k + \epsilon \dot{\widetilde{\boldsymbol{\eta}}}_k^{(e)} \tilde{q}_k + \widetilde{\boldsymbol{\eta}}_k^{(e)} \dot{\tilde{q}}_k + \mathcal{O}(\epsilon^2),\end{aligned} \tag{23}$$

results are obtained for the generalized derivatives $\delta \widetilde{\boldsymbol{\omega}}_k^{(b)}$ and $\delta \widetilde{\mathbf{x}}_k^{(b)}$:

$$\begin{aligned}\delta \widetilde{\boldsymbol{\omega}}_k^{(b)} &= 2 \tilde{q}_k^\dagger \dot{\widetilde{\boldsymbol{\eta}}}_k^{(e)} \tilde{q}_k, \\ \delta \widetilde{\mathbf{x}}_k^{(b)} &= 2 \widetilde{\mathbf{x}}_k^{(b)} \times \tilde{q}_k^\dagger \widetilde{\boldsymbol{\eta}}_k^{(b)} \tilde{q}_k + \frac{1}{h} \tilde{q}_k^\dagger \left(\delta \mathbf{x}_{k+1}^{(e)} - \delta \mathbf{x}_k^{(e)}\right) \tilde{q}_k.\end{aligned} \tag{24}$$

This process has transformed the problem of estimating $\tilde{q}_k^\epsilon$ into one of estimating $\widetilde{\boldsymbol{\eta}}_k^{(b)}$ and $\widetilde{\boldsymbol{\eta}}_k^{(e)}$, as a function of $\boldsymbol{\eta}_k^{(e)}$ and $\boldsymbol{\eta}_{k+1}^{(e)}$. Now, however, we may use linear approximations, which are known to be first order in $h$:

$$\widetilde{\boldsymbol{\eta}}_k^{(e)} = \frac{1}{2}\left(\boldsymbol{\eta}_{k+1}^{(e)} + \boldsymbol{\eta}_k^{(e)}\right), \qquad \dot{\widetilde{\boldsymbol{\eta}}}_k^{(e)} = \frac{1}{h}\left(\boldsymbol{\eta}_{k+1}^{(e)} - \boldsymbol{\eta}_k^{(e)}\right). \tag{25}$$

Formally, we have approximated the interpolation of a quaternion with the interpolation of the quaternion's generators, *cf.* [36]. This leads to estimates for the variational derivatives, as:

$$\begin{aligned}\delta \widetilde{\boldsymbol{\omega}}_k^{(b)} &= \frac{2}{h} \tilde{q}_k^\dagger \left(\boldsymbol{\eta}_{k+1}^{(e)} - \boldsymbol{\eta}_k^{(e)}\right) \tilde{q}_k, \\ \delta \widetilde{\mathbf{x}}_k^{(b)} &= \widetilde{\mathbf{x}}_k^{(b)} \times \tilde{q}_k^\dagger \left(\boldsymbol{\eta}_{k+1}^{(e)} + \boldsymbol{\eta}_k^{(e)}\right) \tilde{q}_k + \frac{1}{h} \tilde{q}_k^\dagger \left(\delta \mathbf{x}_{k+1}^{(e)} - \delta \mathbf{x}_k^{(e)}\right) \tilde{q}_k,\end{aligned} \tag{26}$$

and thus the integrator inter-step equations, in translation and rotation respectively:

$$\begin{aligned}\tilde{q}_k \widetilde{\mathbf{D}}_{1,k} \tilde{q}_k^\dagger &= \tilde{q}_{k-1} \widetilde{\mathbf{D}}_{1,k-1} \tilde{q}_{k-1}^\dagger + h \widetilde{\mathbf{F}}_k^{(e)}, \\ \tilde{q}_k \left(\widetilde{\mathbf{D}}_{2,k} + h \widetilde{\mathbf{x}}_k^{(b)} \times \widetilde{\mathbf{D}}_{1,k}\right) \tilde{q}_k^\dagger &= \tilde{q}_{k-1} \left(\widetilde{\mathbf{D}}_{2,k-1} - h \widetilde{\mathbf{x}}_{k-1}^{(b)} \times \widetilde{\mathbf{D}}_{1,k-1}\right) \tilde{q}_{k-1}^\dagger + \widetilde{\boldsymbol{\tau}}_k^{(b)},\end{aligned} \tag{27}$$

which shows notable similar format to Eq. 15. The midpoint quaternion $\tilde{q}_k$ contains a dependency on $q_{k+1}$ via SLERP interpolation (Eq. 20), and thereby on $\widetilde{\boldsymbol{\omega}}_k^{(b)}$. Assuming constant angular velocity over each step, $\boldsymbol{\omega}^{(b)} = \widetilde{\boldsymbol{\omega}}_k^{(b)}$, we may estimate $q_{k+1}$ as:

$$q_{k+1} = q_k \exp\left(\frac{1}{2} \widetilde{\boldsymbol{\omega}}_k^{(b)}\right). \tag{28}$$

The inter-step equations, Eq. 27, are solved via a Newton's iteration. This completes the derivation of the midpoint QVI.

## 4 NUMERICAL TESTS OF INERTIAL MANEUVERING

### 4.1. Integrator conservation properties under non-morphing conditions

As an initial test of the conservation properties of our integrators, we first study their behavior in a UAV model without wing morphing. Note that, while this non-morphing condition corresponds to the simulation of a single rigid body, it still contains translation-rotation coupling because of the use of an arbitrary body-fixed reference point for the UAV dynamics [10,11]. Under non-morphing conditions, our biomimetic UAV is be described by the constant coefficients [10]:

$$a_{xx} = 4 \text{ kg}, \quad \mathbf{a}_x = \mathbf{a}_\omega = [0\ 0\ 0]^T,$$

$$A_{x\omega} = \begin{bmatrix} & 0.0400 & \\ -0.0400 & & 6.350 \\ & -6.350 & \end{bmatrix} \text{ kg m},$$

$$A_{\omega\omega} = \begin{bmatrix} 0.2342 & & -6.4761 \times 10^{-5} \\ & 3.0539 & \\ -6.4761 \times 10^{-5} & & 3.2699 \end{bmatrix} \text{ kg m}^2. \tag{28}$$

Note that aerodynamic and gravitational forces are excluded, and the system starts with $\boldsymbol{\omega}_k^{(b)} = [1, 1, 1]^T$. For integrator validation, we can also estimate the physical (decoupled) translational and angular momenta of the UAV in postprocessing:

$$\begin{aligned} \mathbf{P}_{\mathbf{x},k} &= m_{\text{tot}} \dot{\mathbf{x}}_{\text{c.o.m.},k}^{(e)}, \\ \mathbf{P}_{\boldsymbol{\omega},k} &= q_k I_{\text{c.o.m.},k}^{(b)} \boldsymbol{\omega}_k^{(b)} q_k^\dagger, \end{aligned} \tag{29}$$

where $m_{\text{tot}}$ is the total UAV mass, $\dot{\mathbf{x}}_{\text{c.o.m.}}^{(e)}$ the velocity of the UAV center of mass in the earth frame, and $I_{\text{c.o.m.}}^{(b)}$ the total UAV rotational inertia about the center of mass, resolved in the body-fixed frame. Being decoupled, these momenta are independently conserved, and so can be used to estimate decoupled translational and angular momentum conservation errors:

Translational momentum error: $\quad e_{\mathbf{x},k} = \max_{i \leq k}(\|\mathbf{P}_{\mathbf{x},i} - \mathbf{P}_{\mathbf{x},1}\|/\|\mathbf{P}_{\mathbf{x},1}\|)$

Angular momentum error: $\quad e_{\boldsymbol{\omega},k} = \max_{i \leq k}(\|\mathbf{P}_{\boldsymbol{\omega},i} - \mathbf{P}_{\boldsymbol{\omega},1}\|/\|\mathbf{P}_{\boldsymbol{\omega},1}\|)$ (30)

In addition, the kinetic energy conservation error can be estimated directly from Eq. 1:

Kinetic energy error: $\quad e_{T,k} = \max_{i \leq k}(|T_i - T_1|/T_1).$ (31)

Fig. 2A shows results from the midpoint and left-rectangle QVIs, as well as an adaptive Runge-Kutta (RK) integrator in Euler angles with pole-switching, developed previously for this biomimetic UAV [10,11]. Fig. 2B illustrates the conservation behavior of these integrators in energy and momentum. Several points may be noted. Both QVIs are effectively exact in translational momentum conservation – errors remain below machine precision, $10^{-13}$ – as a result of their symmetry in translational inter-step dynamics (Eq. 15, 27). However, the midpoint QVI significantly outperforms the left-rectangle QVI in energy and rotational momentum conservation: errors are consistently lower by a factor of $10^2$. An additional notable feature of the midpoint QVI's angular momentum conservation is that the angular momentum error is oscillatory, with minimal long-term drift: this is illustrated in Fig. 2C. The purely oscillatory nature of this error suggests that there may be a method of accounting for this dynamical effect exactly (*e.g.*, via more detailed analysis of Eq. 30) and thereby reducing all momentum errors to the level of machine precision. It also provides assurance regarding the long-term behavior of the midpoint QVI: the only conserved property which shows steady deviation is the kinetic energy.

**4.2. Integrator performance under inertial maneuvering behavior**

Finally, as a test of the midpoint QVI for true inertial maneuvering, Fig. 3 shows a maneuver arising from an oscillatory pattern of UAV wingbeat motion. The symmetric wing morphing inputs are dihedral $\theta_w(t) = \sin(t)$ rad and incidence $\phi_w(t) = -0.5\cos(t)$ rad. With wing morphing active, the system coefficients are now time-varying. Following per Bergou *et al.* [4], we simulate a bio-inspired pitching maneuver, including aerodynamic forces, but neglecting gravitational ones. This motion could be of utility in the design of UAV vertical landing maneuvers, analogous to those carried out by bats [4]. Fig. 3 shows the kinematic history of the UAV over the simulation, alongside a maneuver rendering. Results are presented for the QVI timestep, $h = h_{\text{RK}} = 183$ ms, where $h_{\text{RK}}$ is the mean step size of the RK integrator over the simulation.

The midpoint QVI performs excellently, demonstrating the effectiveness of the QVI for practical inertial maneuvering simulation. It additionally demonstrates the differences between the QVI and the RK integrator in terms of the suitable step size. In this inertial maneuvering simulation, the QVI is capable of matching and exceeding the adaptive RK step size; whereas in the free rotation and translation test case these integrators require smaller step sizes. Further numerical analysis techniques may be able to shed light on these effects, but given the complexity of the integrators system-specific testing is likely to be a more practical approach for assessing integrator suitability. Overall, this midpoint QVI is well-suited to the analysis of complex inertial maneuvering in UAVs; and shows potential for applications in other systems showing complex dynamics.

## 5. CONCLUSION

In this work, we developed a pair of QVIs for simulating the strongly coupled dynamics that govern inertial maneuvering in biomimetic UAV. Formulating the inertial maneuvering dynamics in terms of a general quaternion-parameterized system with strong coupling, we developed integrators under two variational integration conditions: left-rectangle and midpoint variational integration. Resolving aspects such as the appropriate definitions of quaternion and perturbation interpolation, we applied these two integrators to numerical tests of inertial maneuvering behavior in a biomimetic UAV showing complex coupling behavior, and studied their behavior. The midpoint QVI is best suited to the analysis of inertial maneuvering behavior: with the canonical momenta of the system varying due to coupling, a midpoint approximation of these momenta is significantly more precise than a conventional left-rectangle approximation. The midpoint QVI performs excellently in energy and momentum conservation, and is a candidate for precise and singularity-free simulation of practical inertial maneuvering behavior in biomimetic UAVs and other complex systems.

**FUNDING**

This work was supported by the Cambridge Commonwealth Trust.

**NOMENCLATURE**

| | |
|---|---|
| $\boldsymbol{\omega}$ | fuselage angular velocity vector, rad/s |
| $\dot{\mathbf{x}}$ | fuselage velocity vector, m/s |
| $q$ | orientation quaternion |
| $\mathbf{v}$ | set of system control and structural parameters |
| $t$ | time, s |
| $h$ | step size, s |
| $T$ | kinetic energy, J |
| A, **a**, a | multibody model coefficient |
| $\mathbf{p}$ | metric of canonical momentum |
| $\mathbf{P}$ | metric of physical momentum |
| $\mathbf{Q}$ | generalized force |
| $\mathbf{r}$ | generalized coordinate |
| I | rotational moment of inertia, kg m$^2$ |
| $\delta$ | variational derivative |
| $\boldsymbol{\eta}$ | variational perturbation direction |
| $\epsilon$ | infinitesimal for perturbation |
| $\otimes$ | quaternion multiplication |
| $\boldsymbol{\theta}$ | quaternion mapping parameter |
| $(\tilde{\phantom{x}})$ | evaluation at step midpoint |
| $\|\cdot\|$ | Euclidean norm |
| $\mathcal{O}(n)$ | order of magnitude of $n$ |

**Superscripts**

| | |
|---|---|
| $T$ | matrix transpose |
| † | quaternion transpose |
| $(b)$ | resolution in body-fixed reference frame |
| $(e)$ | resolution in earth reference frame |

**Subscripts**

| | |
|---|---|
| $k$ | integration step |

# REFERENCES

[1] Kane, T. R., and Scher, M. P., 1969, "A Dynamical Explanation of the Falling Cat Phenomenon," Int. J. Solids Struct., **5**(7), pp. 663–670. DOI: 10.1016/0020-7683(69)90086-9.

[2] Siddall, R., Ibanez, V., Byrnes, G., Full, R. J., and Jusufi, A., 2021, "Mechanisms for Mid-Air Reorientation Using Tail Rotation in Gliding Geckos," Integr. Comp. Biol., **61**(2), pp. 478–490. DOI: 10.1093/icb/icab132.

[3] Jusufi, A., Zeng, Y., Full, R. J., and Dudley, R., 2011, "Aerial Righting Reflexes in Flightless Animals," Integr. Comp. Biol., **51**(6), pp. 937–943. DOI: 10.1093/icb/icr114.

[4] Bergou, A. J., Swartz, S. M., Vejdani, H., Riskin, D. K., Reimnitz, L., Taubin, G., and Breuer, K. S., 2015, "Falling with Style: Bats Perform Complex Aerial Rotations by Adjusting Wing Inertia," PLOS Biol., **13**(11), p. e1002297. DOI: 10.1371/journal.pbio.1002297.

[5] Hedrick, T. L., Usherwood, J. R., and Biewener, A. A., 2007, "Low Speed Maneuvering Flight of the Rose-Breasted Cockatoo (Eolophus Roseicapillus). II. Inertial and Aerodynamic Reorientation," J. Exp. Biol., **210**(11), pp. 1912–1924. DOI: 10.1242/jeb.002063.

[6] Yeaton, I. J., Ross, S. D., Baumgardner, G. A., and Socha, J. J., 2020, "Undulation Enables Gliding in Flying Snakes," Nat. Phys., **16**(9), pp. 974–982. DOI: 10.1038/s41567-020-0935-4.

[7] Charlet, M., and Gosselin, C., 2022, "Reorientation of Free-Falling Legged Robots," ASME Open J. Eng., **1**, p. 011009. DOI: 10.1115/1.4053897.

[8] Liu, Y., and Ben-Tzvi, P., 2020, "Design, Analysis, and Integration of a New Two-Degree-of-Freedom Articulated Multi-Link Robotic Tail Mechanism," J. Mech. Robot., **12**(2), p. 021101. DOI: 10.1115/1.4045842.

[9] Rone, W. S., Saab, W., Kumar, A., and Ben-Tzvi, P., 2019, "Controller Design, Analysis, and Experimental Validation of a Robotic Serpentine Tail to Maneuver and Stabilize a Quadrupedal Robot," J. Dyn. Syst. Meas. Control, **141**(8), p. 081002. DOI: 10.1115/1.4042948.

[10] Pons, A., 2019, "Supermanoeuvrability in a Biomimetic Morphing-Wing Aircraft," PhD Thesis, University of Cambridge.

[11] Pons, A., and Cirak, F., 2021, "Multi-Axis Nose-Pointing-and-Shooting in a Biomimetic Morphing-Wing Aircraft," J. Guid. Control Dyn., article in advance. DOI: 10.2514/1.G006381.

[12] Drury, R. G., and Whidborne, J. F., 2009, "Quaternion-Based Inverse Dynamics Model for Expressing Aerobatic Aircraft Trajectories," J. Guid. Control Dyn., **32**(4), pp. 1388–1391. DOI: 10.2514/1.42883.

[13] Terze, Z., Zlatar, D., Vrdoljak, M., and Pandža, V., 2017, "Lie Group Forward Dynamics of Fixed-Wing Aircraft With Singularity-Free Attitude Reconstruction on SO(3)," J. Comput. Nonlinear Dyn., **12**(2), p. 021009. DOI: 10.1115/1.4034398.

[14] Kuipers, J. B., 1999, *Quaternions and Rotation Sequences*, Princeton University Press, Princeton, NJ.

[15] Leitz, T., and Leyendecker, S., 2018, "Galerkin Lie-Group Variational Integrators Based on Unit Quaternion Interpolation," Comput. Methods Appl. Mech. Eng., **338**, pp. 333–361. DOI: 10.1016/j.cma.2018.04.022.

[16] Andrle, M. S., and Crassidis, J. L., 2013, "Geometric Integration of Quaternions," J. Guid. Control Dyn., **36**(6), pp. 1762–1767. DOI: 10.2514/1.58558.

[17] Goodarzi, F. A., Lee, D., and Lee, T., 2015, "Geometric Adaptive Tracking Control of a Quadrotor Unmanned Aerial Vehicle on SE(3) for Agile Maneuvers," J. Dyn. Syst. Meas. Control, **137**(9), p. 091007. DOI: 10.1115/1.4030419.

[18] Manchester, Z. R., and Peck, M. A., 2016, "Quaternion Variational Integrators for Spacecraft Dynamics," J. Guid. Control Dyn., **39**(1), pp. 69–76. DOI: 10.2514/1.G001176.

[19] Afonso Silva, F. F., José Quiroz-Omaña, J., and Vilhena Adorno, B., 2022, "Dynamics of Mobile Manipulators Using Dual Quaternion Algebra," J. Mech. Robot., **14**(6), p. 061005. DOI: 10.1115/1.4054320.

[20] Ruggiero, A., and Sicilia, A., 2022, "A Musculoskeletal Multibody Algorithm Based on a Novel Rheonomic Constraints Definition Applied to the Lower Limb," J. Biomech. Eng., **144**(8), p. 081010. DOI: 10.1115/1.4053874.

[21] Xu, D., Jahanchahi, C., Took, C. C., and Mandic, D. P., 2015, "Enabling Quaternion Derivatives: The Generalized HR Calculus," R. Soc. Open Sci., **2**(8), p. 150255. DOI: 10.1098/rsos.150255.

[22] Müller, A., Terze, Z., and Pandza, V., 2016, "A Non-Redundant Formulation for the Dynamics Simulation of Multibody Systems in Terms of Unit Dual Quaternions," Proceedings of the IDETC/CIE, Charlotte, NC, 2016. ASME Paper No. V006T09A011. DOI: 10.1115/DETC2016-60191.

[23] Müller, A., and Terze, Z., 2016, "Geometric Methods and Formulations in Computational Multibody System Dynamics," Acta Mech., **227**(12), pp. 3327–3350. DOI: 10.1007/s00707-016-1760-9.

[24] Crouch, P. E., and Grossman, R., 1993, "Numerical Integration of Ordinary Differential Equations on Manifolds," J. Nonlinear Sci., **3**(1), pp. 1–33. DOI: 10.1007/BF02429858.

[25] Munthe-Kaas, H., 1998, "Runge-Kutta Methods on Lie Groups," BIT Numer. Math., **38**(1), pp. 92–111. DOI: 10.1007/BF02510919.

[26] Sveier, A., Sjøberg, A. M., and Egeland, O., 2019, "Applied Runge–Kutta–Munthe-Kaas Integration for the Quaternion Kinematics," J. Guid. Control Dyn., **42**(12), pp. 2747–2754. DOI: 10.2514/1.G004578.

[27] Leitz, T., Sato Martín de Almagro, R. T., and Leyendecker, S., 2021, "Multisymplectic Galerkin Lie Group Variational Integrators for Geometrically Exact Beam Dynamics Based on Unit Dual Quaternion Interpolation — No Shear Locking," Comput. Methods Appl. Mech. Eng., **374**, p. 113475. DOI: 10.1016/j.cma.2020.113475.

[28] Lee, T., Leok, M., and McClamroch, N. H., 2007, "Lie Group Variational Integrators for the Full Body Problem," Comput. Methods Appl. Mech. Eng., **196**(29–30), pp. 2907–2924. DOI: 10.1016/j.cma.2007.01.017.

[29] Saccon, A., 2009, "Midpoint Rule for Variational Integrators on Lie Groups," Int. J. Numer. Methods Eng., **78**(11), pp. 1345–1364. DOI: 10.1002/nme.2541.

[30] Marsden, J. E., and West, M., 2001, "Discrete Mechanics and Variational Integrators," Acta Numer. 2001, **10**, pp. 357–514. DOI: 10.1017/S096249290100006X.

[31] Sola, J., 2016, *Quaternion Kinematics for the Error-State KF*, Technical Report IRI-TR-16-02, Universitat Politecnica de Catalunya, Barcelona, Spain.

[32] Lee, T., Leok, M., and McClamroch, N. H., 2018, *Global Formulations of Lagrangian and Hamiltonian Dynamics on Manifolds*, Springer, Cham, Switzerland.

[33] Kibble, T. W. B., and Berkshire, F. H., 2004, *Classical Mechanics*, Imperial College Press, London, UK.

[34] Dam, E. B., Koch, M., and Lillholm, M., 1998, *Quaternions, Interpolation and Animation*, Technical Report DIKU-TR-98/5, University of Copenhagen, Copenhagen, Denmark.

[35] Markley, F. L., Cheng, Y., Crassidis, J. L., and Oshman, Y., 2007, "Averaging Quaternions," J. Guid. Control Dyn., **30**(4), pp. 1193–1197. DOI: 10.2514/1.28949.

[36] Boyle, M., 2017, "The Integration of Angular Velocity," Adv. Appl. Clifford Algebr., **27**(3), pp. 2345–2374. DOI: 10.1007/s00006-017-0793-z.

## Figure Captions List

Fig. 1   Inertial maneuvering in a biomimetic morphing-wing UAV. (**A**) UAV system, equipped with wing incidence, dihedral and sweep angle control. (**B**) Hypothetical application of biomimetic inertial maneuvering, based on maneuvers carried out by bats [4]: wing inertial control enables take-off from, and landing on, an inverted surface.

Fig. 2   Integration results for a coupled rigid body: comparison between left-rectangle QVI ($h = 0.01$), midpoint QVI ($h = 0.01$), and adaptive RK integrator in Euler angles (average step size $h_{RK} = 0.095$). (**A**) Sample trajectories in angular velocity ($\boldsymbol{\omega}^{(b)}$) and translational velocity ($\dot{\mathbf{x}}^{(b)}$). (**B**) Conservation errors in kinetic energy, translational momentum, and rotational momentum. (**C**) Rotational momentum error history for the midpoint QVI integrator.

Fig. 3   Integration results for the midpoint QVI applied to a biomimetic pitching maneuver. (**A**) Integrator results for the QVI and the adaptive RK integrator in Euler angles: in pitch rate, pitch angle, body velocity and center of mass (C.O.M.) location. (**B**) Rendering of maneuver, for both integrators.to separate the multiple sentences.

# Figures

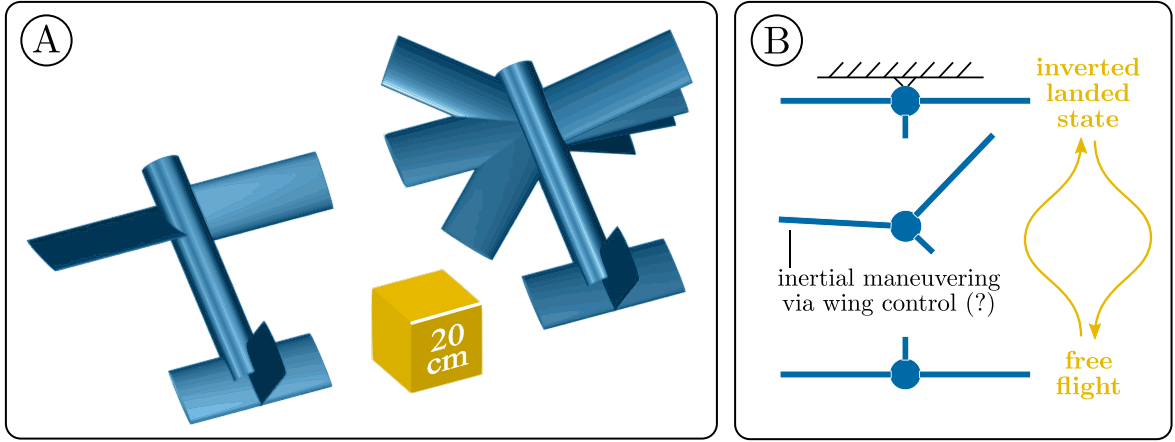

*Fig. 1*

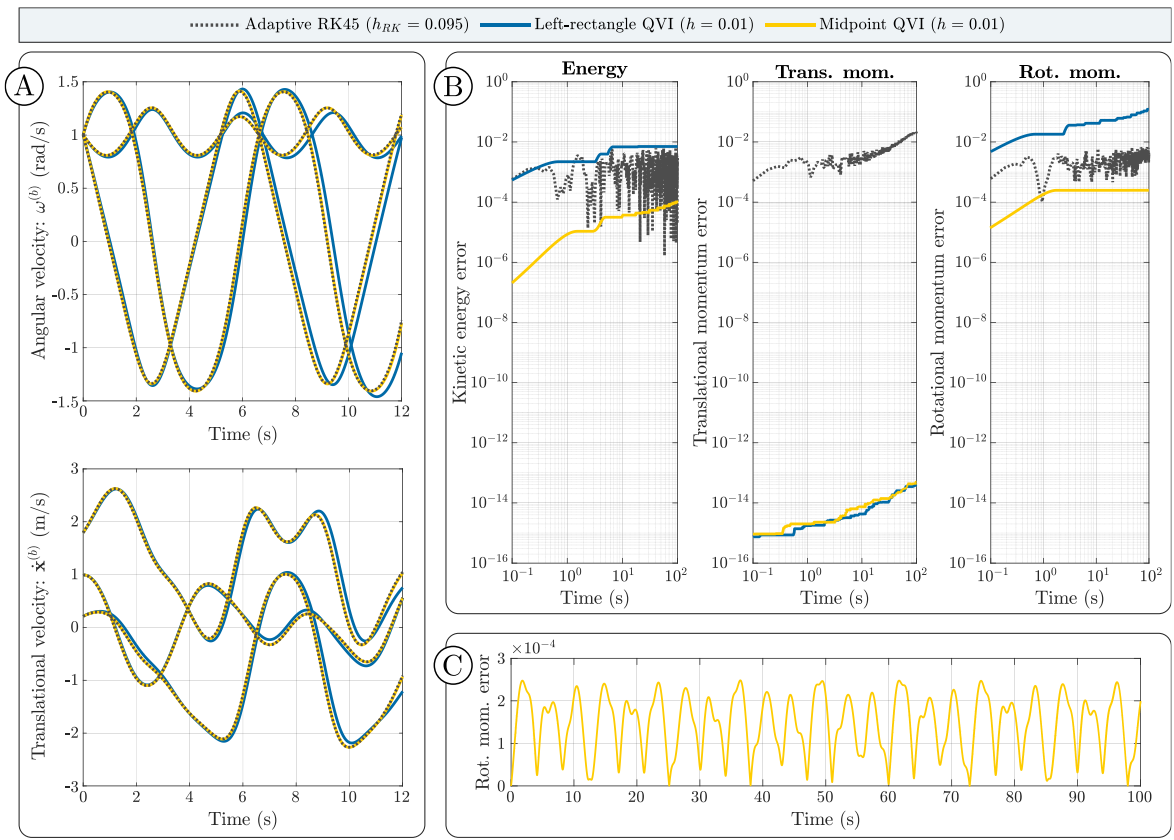

*Fig. 2*

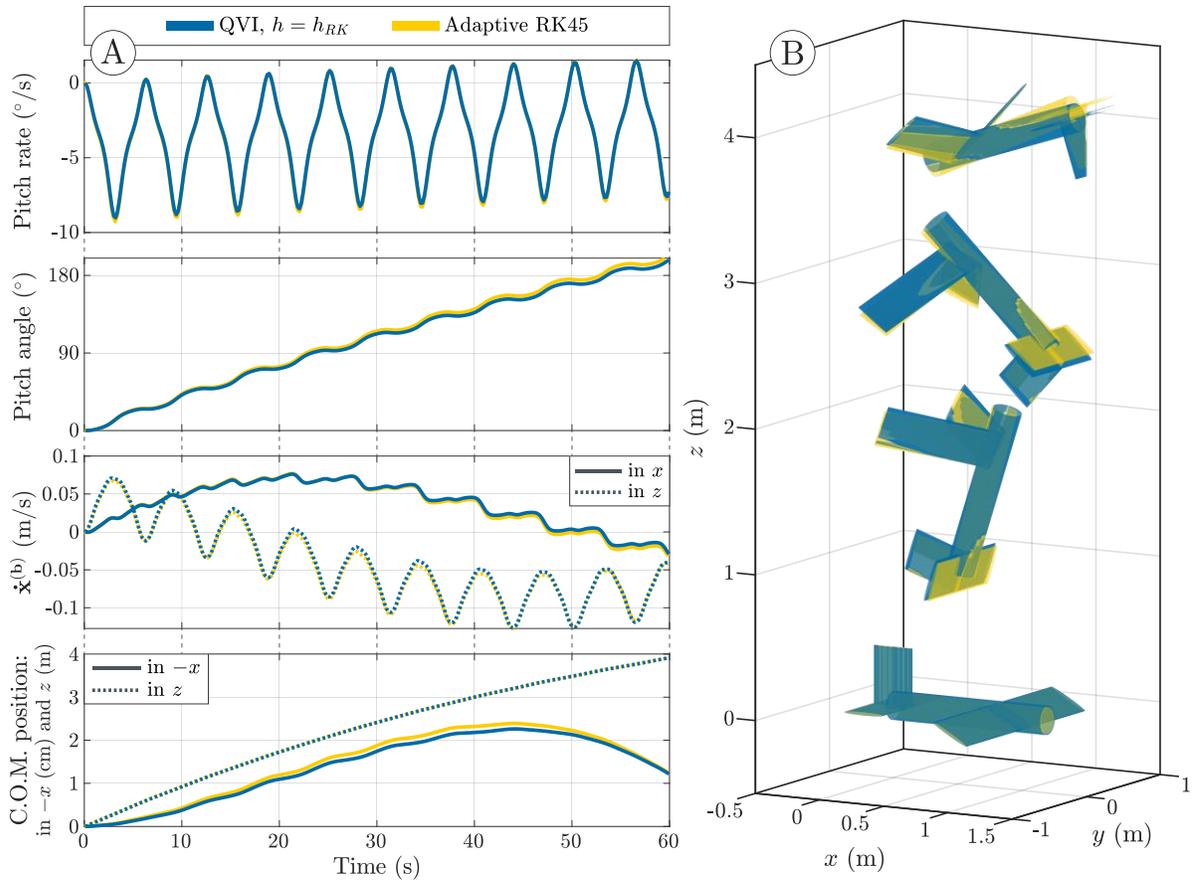

*Fig. 3*